\documentclass{ieeeconf}
\usepackage{spconf,amsmath,graphicx}
\usepackage{graphics} 
\usepackage{epsfig} 
\usepackage{amssymb}  
\usepackage{multirow}
\usepackage{array}
\usepackage{float}
\usepackage{booktabs,subcaption,amsfonts,dcolumn}
\usepackage{algorithm}
\usepackage{algorithmic}
\usepackage{caption}
\usepackage{colortbl}
\usepackage{multicol}
\usepackage{adjustbox}
\usepackage{tabularx}
\usepackage{booktabs}
\usepackage[obeyspaces]{url}

\usepackage{mars}

\newcolumntype{Y}{>{\raggedright\arraybackslash}X}
\newcolumntype{b}{Y}
\newcolumntype{z}{>{\hsize=.55\hsize}Y}
\newcolumntype{s}{>{\hsize=.15\hsize}Y}

\title{\LARGE \bf
	Improved Fourier Mellin Invariant for\\ Robust Rotation Estimation with Omni-cameras
}

\author{Qingwen Xu$^{1}$, Arturo Gomez Chavez$^{2}$, Heiko B\"ulow$^{2}$, Andreas Birk$^{2}$ and S\"oren Schwertfeger$^{1}$
	\thanks{$^{1}$Authors are with School of Information Science Technology of ShanghaiTech University
		{\tt\small <xuqw, soerensch>@shanghaitech.edu.cn}}%
	\thanks{$^{2}$Authors are with the Dept. of Electrical Engineering \& Computer Science, Jacobs University, Bremen, Germany.
		{\tt\small <a.gomezchavez, h.buelow, a.birk>@jacobs-university.de}}%
}

\title{Improved Fourier Mellin Invariant for\\ Robust Rotation Estimation with Omni-cameras}
\name{{Qingwen Xu$^{1}$ \quad Arturo Gomez Chavez$^{2}$ \quad Heiko B\"ulow$^{2}$ \quad Andreas Birk$^{2}$ \quad S\"oren Schwertfeger$^{1}$} \thanks{This work is supported by DAAD Scholarship, Funding No. 57378443. A pre-print version of this paper can be found on Arxiv: https://arxiv.org/abs/1811.05306.}}
\address{$^{1}$Authors are with School of Information Science Technology of ShanghaiTech University \\ $^{2}$Authors are with the Robotics Group, Computer Science \& Electrical Engineering,\\Jacobs University Bremen, Germany}
\begin{document}
	%
	
	
	\marsPublishedIn{Published with:} 	
	
	\marsVenue{IEEE International Conference on Image Processing (ICIP) 2019}
	
	\marsYear{2019}
	
	\marsPlainAutors{\large{Qingwen Xu$^{1}$, Arturo Gomez Chavez$^{2}$, Heiko B\"ulow$^{2}$, Andreas Birk$^{2}$ and S\"oren Schwertfeger$^{1}$}}

	
	\marsMakeCitation{Improved Fourier Mellin Invariant for Robust Rotation Estimation with Omni-cameras}{IEEE Press}
	
	\marsDOI{\url{https://dx.doi.org/10.1109/ICIP.2019.8802933}}
	
	\marsIEEE{}
	
	
	\makeMARStitle
	
	%
	%
	%
	\maketitle
	\begin{abstract}
		Spectral methods such as the improved Fourier Mellin Invariant (iFMI) transform have proved to be faster, more robust and accurate than feature based methods on image registration. 
		However, iFMI is restricted to work only when the camera moves in 2D space and has not been applied on omni-cameras images so far. 
		In this work, we extend the iFMI method and apply a motion model to estimate an omni-camera's pose when it moves in 3D space.  
		In the experiment section, we compare the extended iFMI method against ORB and AKAZE feature based approaches on three datasets, showing different types of environments: 
		office, lawn and urban scenery (\emph{MPI-omni} dataset).
		The results show that our method reduces the error of the camera pose estimation two to three times with respect to the feature registration techniques, while offering lower processing times. 
	\end{abstract}
	\vspace{-0.15cm}
	\section{Introduction}
	\label{sec:intro}
	\vspace{-0.15cm}
	Fourier-Mellin transformation is used in pattern recognition and image registration due to its advantage in accuracy and robustness. Chen et al. used the Fourier-Mellin invariant (FMI) descriptor and symmetric phase-only matched filtering to calculate rotation, scaling and translation in \cite{chen1994symmetric}. B\"ulow et al. improve the FMI (iFMI) descriptor to stitch images faster and more accurately in \cite{bulow2009fast}. 
	Then, the work in \cite{kazik2011visual} proposed a FMI based visual odometry method for a ground robot, which proved to be more accurate than optical flow based method.
	However, the FMI and iFMI methods are only applicable to 2D motion when it comes to visual odometry.
	Kaustubh et al. proposed an iFMI extension to calculate the camera tilt \cite{pathak2013robust}, including pitch and roll for forward looking cameras, but estimating yaw remains a challenge.
	
	Our work extends the iFMI method to estimate the orientation of the omni-camera when it moves in 3D space. To overcome the issue that yaw cannot be calculated through the iFMI method for forward looking cameras, we exploit the property of the omni-camera to convert the 3D rotation estimation to a 2D image motion computation.  
	We compare our method to feature-based approaches in different indoor and outdoor environments. Evaluations are made regarding computation time, robustness and accuracy. 
	ORB\cite{rublee2011orb} and AKAZE\cite{alcantarilla2011akaze} were chosen as baselines, because ORB has the fastest computation time while AKAZE is designed to handle large image distortions, which are commonly present in omni-images according to \cite{tareen2018comparative}. 
	In summary, our main contributions include:
	
	\begin{itemize}
		\setlength{\itemsep}{0pt}
		\setlength{\parsep}{0pt}
		\setlength{\parskip}{0pt}
		\item extending the iFMI method to estimate motion between omni-images; hence
		\item allowing 3D registration instead of 2D for iFMI;
		\item proposing a motion model based on sub-image patches to compensate for omni-images non-linear distortions;
		\item providing baseline comparisons against commonly used registration feature-based methods.
	\end{itemize}
	
	
	\vspace{-0.5cm}
	\section{Omni-Camera Model}
	\label{sec: ocam_model}
\begin{figure}[tbp]
		\centering
		\includegraphics[width=\linewidth]{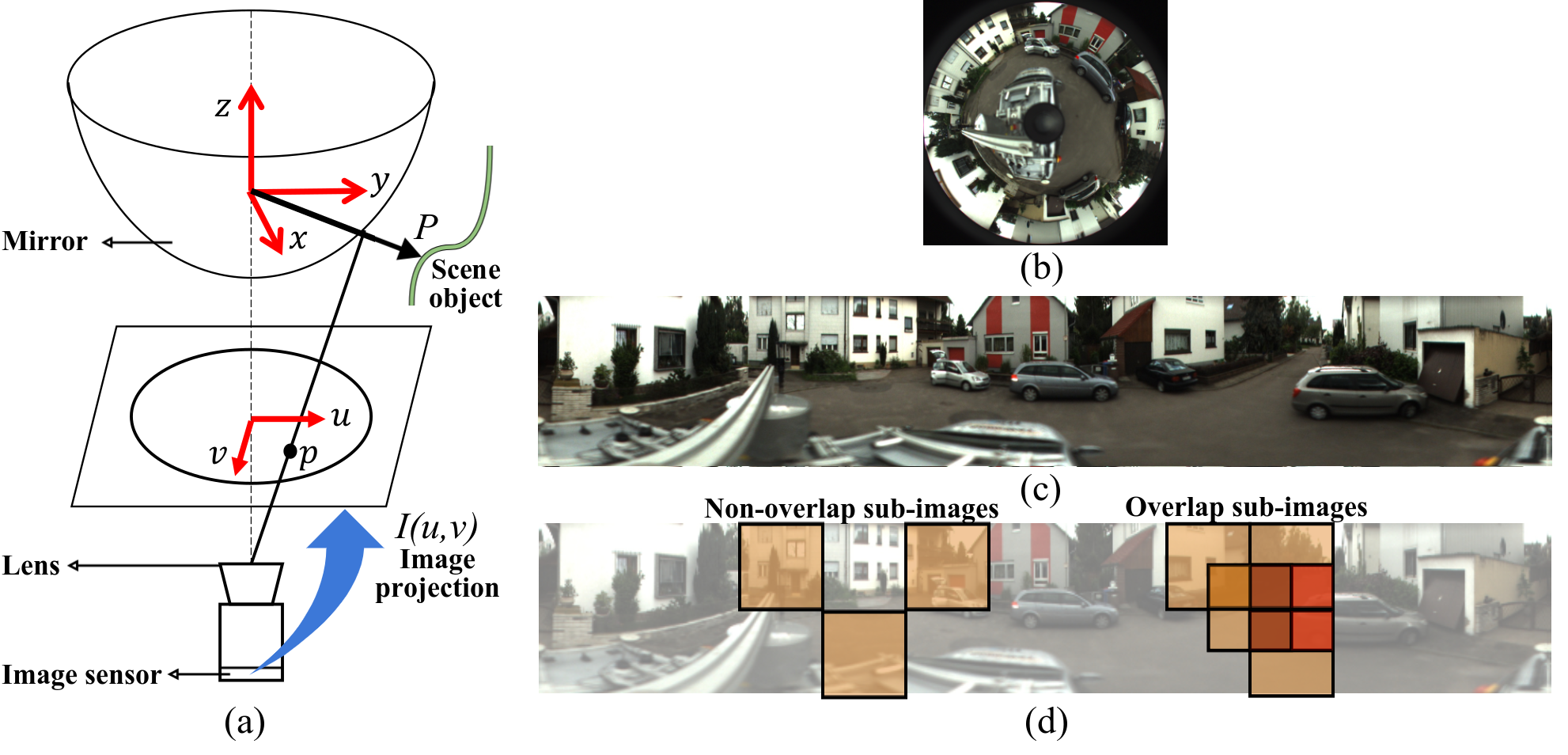}
		\caption{(a) Catadioptric Omni-directional Camera Model \cite{scaramuzza2006flexible}. (b) Omnidirectional image $I(u,v)$. (c) Panorama image obtained from omni-image. (d) Non-overlapping and overlapping sub-image extraction.}
		\label{fig:cameramodel}
		
	\end{figure}

	\vspace{-0.15cm}
	In recent years, there has been a large amount of publications on omni-camera models. 
	To represent the relation between image pixels and camera rays (Fig.~\ref{fig:cameramodel}a), we exploit the model in \cite{scaramuzza2006flexible}, since it is independent of the type of camera sensors and easy to calibrate with the provided toolbox. 
	Two assumptions are taken into consideration: the camera center and omni-lens are aligned and the omni-lens rotates symmetrically. 
	
	Suppose the coordinate of a pixel $p$ in the omni image $I_o$ is $(u,v)$, when the origin is in the center of $I_o$ and its corresponding camera ray $\mathbf{P}$, defined by 
	\begin{equation}
	\small
	P={\left[ \begin{array}{c}
		x\\
		y\\
		z
		\end{array} 
		\right]} = {\left[ \begin{array}{c}
		\alpha u \\
		\alpha v \\
		f(u,v)
		\end{array}
		\right]} => {\left[ \begin{array}{c}
		u \\
		v \\
		f(u,v) 
		\end{array}
		\right]} = \pi^{-1}(p) \;,
	\label{eq:cameramodel1}
	\end{equation}
	show the direction of the corresponding real world scene point from this pixel.  
	In this case, since the mirror is rotationally symmetric, $f(u,v)$ depends only on the distance $\rho = \sqrt{u^2+v^2}$ of a point to the image center. 
	$f(\rho)$ is modeled as a high degree polynomial given by
	\begin{equation}
	\small
	f(u,v) = f(\rho) = \alpha_0 + \alpha_1 \rho + \alpha_2 \rho ^2 + \alpha_3 \rho ^3 + ... 
	\label{eq:cameramodel2}
	\end{equation}
	to represent different types of lenses.
	Finally, to account for possible errors in the hypothesis about the camera-lens alignment and coordinates transformation of the omni-image $I_o$, an extra affine transformation is used as follows,
	\begin{equation}
	\small
	{\left[ \begin{array}{c}
		u\ast \\
		v\ast
		\end{array} 
		\right]} = {\left[ \begin{array}{cc}
		c & d \\
		e & 1 
		\end{array}
		\right]}{\left[ \begin{array}{c}
		u \\
		v 
		\end{array}
		\right]} + {\left[\begin{array}{c}
		x_c' \\
		y_c'
		\end{array}\right]}\;.
	\label{eq:cameramodel3}
	\end{equation}
	where $(x_c', y_c')$ is the center of the omni-image and $(u\ast, v\ast)$ is the coordinate of $p$ when the origin is in the upper left corner of $I_o$. 
	
	
	\vspace{-0.15cm}
	\section{Extending the iFMI Method for Omni-Cameras in 3D space}
	\label{sec:Extend_iFMI}
	\vspace{-0.15cm}
	
	The first step is to convert the omni-image into a panorama image by representing each pixel $(u,v)$ in polar coordinates and mapping these into the Cartesian plane, where each pixel is denoted as $(u',v')$.
	This is done to avoid determining the type of image transformation, affine or projective, between frames by replacing with pixels' motion when the camera moves in 3D space.
	Next, to ensure consistent motion among all image pixels when moving, we extract small local regions or sub-images $a_i$.
	Consistent motion among sub-image pixels can be ensured if the panorama image is divided in sufficiently small regions.  
	Then, the motion between two sub-images $({}^ka_i,{}^{k+1}a_i)$ from consecutive frames $k$ and $k+1$ can be estimated using the iFMI method.

	\vspace{-0.15cm}
	\subsection{Sub-images Extraction}
	\label{ssec:sub-image}
	The panorama image is divided into multiple sub-images with size $N_a\times N_a$ in both overlapping and non-overlapping fashion as shown in Fig.~\ref{fig:cameramodel}d. The best sub-image extraction parameters were chosen after comparing the computation time and motion estimation error among sub-images with different size and overlapping percentage. Sub-images with size $N_a$ equals to $\approx 10\%$ of the panorama image width and no overlap is used in this work according to our experiments. According experiments are not shown due to space constraints. 
	
	\vspace{-0.15cm}
	\subsection{Motion Estimation between Sub-images} 
	First, we find corresponding sub-image pairs from consecutive panorama frames. Assume these two sub-images sets are $^1\mathbb{A} = \{^1a_1, ^1a_2,...,^1a_m\}$ and $^2\mathbb{A} = \{^2a_1,^2a_2,...,^2a_n\}$, $^1a_i$ and $^2a_i$ are considered as a sub-image pair based on the assumption that images are captured sequentially. 
	
	The motion between the sub-images $^1a_i$ and $^2a_i$ is calculated with the iFMI method, described in detail in~\cite{chen1994symmetric,bulow2009online}.
	To summarize iFMI, image rotation is the same in the spectral magnitude representation $\lvert G \rvert$, and image scaling by $\sigma$ scales $\lvert G \rvert$ by a factor of $\sigma^{-1}$ according to
	\begin{equation}
	\small
	\lvert G(^1a_i) \rvert = \sigma^{-2} \lvert G( \sigma^{-1} \cdot {}^2a_i) \rvert\;.
	\label{eq:spectral_relation}
	\end{equation}  
	We are obtaining the iFMI descriptor $V_G$ by resampling the spectra to polar coordinates. The radial axes are represented logarithmically and then the Fourier transform is computed. Thus Eq.~\eqref{eq:spectral_relation} can be expressed as 
	\begin{equation}
	\small
	V_{G(^1a_i)}= \sigma^{-2} e^{-j2\pi r(\theta,s)} V_{G(^2a_i)}\;,
	\label{eq:fourier-mellin}
	\end{equation} 
	which effectively decouples rotation, scaling and translation. 
	All these transformations are now represented by signal shifts $(\theta,s)$ which can be estimated using Phase-Only Matched Filter. After re-scaling and re-rotating the image, translation between two images can be calculated by matched filter.
	
	\vspace{-0.15cm}
	\section{Motion Model}
	\label{sec: motion_model}
	\vspace{-0.15cm}
	Assume two images $^1I_o$ and $^2I_o$ are captured by the omni-camera in two different poses $^1C$ and $^2C$, so that the motion model describes the transformation ${^2_1T}$ between poses $^1C$ and $^2C$. We construct the motion model into two steps: building the motion flow field and estimating the transformation ${^2_1T}$. 
	
	\vspace{-0.3cm}
	\subsection{Motion Flow Field}
\begin{figure}
		\centering
		\subcaptionbox{First Panorama Image\label{fig:pano right}}{
			\includegraphics[width=0.98\columnwidth]{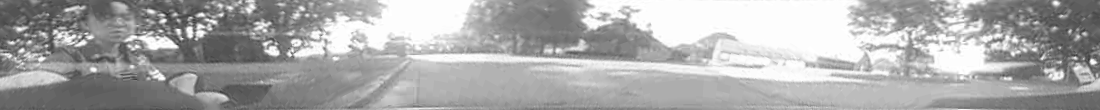} 
		}\\
		\subcaptionbox{Second Panorama Image\label{fig:pano_left}}{
			\includegraphics[width=0.98\columnwidth]{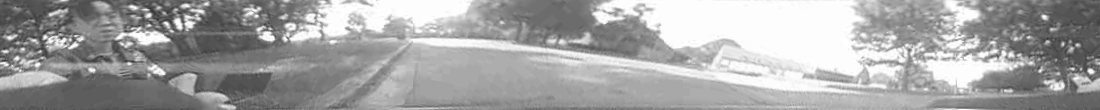} 
		}
		\subcaptionbox{Motion Flow Field\label{fig:motion_flow_field}}{
			\includegraphics[width=0.98\columnwidth]{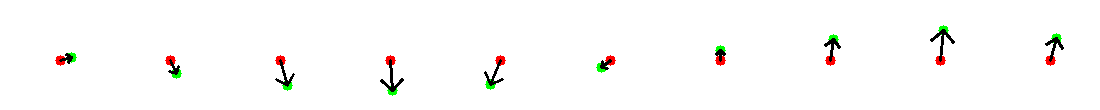} 
		}
		\captionsetup{justification=justified}
		\caption{Motion flow field between panorama images divided in 10 sub-images. The red dot is pixel $^1p_{a_i}$ in the first sub-image while the green is the corresponding pixel $^2p_{a_i}$ in the second one. Arrows represent motion between sub-images (scaled for displaying purposes).}
		\label{fig:motion flow}
	\end{figure}

	To build the motion flow field, corresponding pixel pairs between two panorama images need to be found, i.e. corresponding pixel pairs between sub-image sets $^1\mathbb{A},^2\mathbb{A}$. 
	First, the relative motion $m_i = [s\;\theta\; t_x\; t_y]^\top$ between the corresponding sub-image pair $(^1a_i, ^2a_i)$ is computed through the extended iFMI method in Section \ref{sec:Extend_iFMI}. Then the relation between pixel $^2p_{a_j} = (u'_2,v'_2)$ in $^2a_j$ and its corresponding pixel $^1p_{a_i} = (u'_1,v'_1)$ in $^1a_i$ is represented by
	\begin{equation}
	\small
	\left[\begin{array}{c}
	u'_1 \\
	v'_1
	\end{array}
	\right] = \left[\begin{array}{c}
	u'_2\alpha - v'_2\beta + c_x(1-\alpha) + c_y\beta + t_x \\
	u'_2\beta + v'_2\alpha - c_x\beta + c_y(1-\alpha) + t_y
	\end{array}
	\right]\;,
	\label{eq:motionFlow}
	\end{equation}
	where $\alpha = s\cos\theta$, $\beta = s\sin\theta$ and $(c_x,c_y)$ is the center pixel of the sub-image as discussed in~\cite{szeliski2010computer}. 
	Based on this, all corresponding pixels of sub-image $^2a_j$ can be found in $^1a_j$ via Eq.~\eqref{eq:motionFlow}. 
	Fig.~\ref{fig:motion flow} shows a motion flow field example for several arbitrary pixels.
	
	\vspace{-0.3cm}	
	\subsection{Relative Pose Estimation}
	The transformation ${^2_1T}$ between two camera poses $^1C$ and $^2C$ can be estimated by pixel correspondences found by Equation~\eqref{eq:motionFlow}. 
	First, we use the OCamCalib toolbox \cite{rufli2008automatic,scaramuzza2006toolbox} to calibrate the camera and associate each pixel $(u,v)$ in $^1I_o$ to the undistorted and normalized camera ray $^1P = [u, v,f(u,v)]^T$ in the $^1C$ coordinate frame; following the camera model described in Eq.~\eqref{eq:cameramodel1}. Same procedure is done for the image $^2I_o$. 
	
	Then, based on the pixel pairs $(^1p_{a_i},^2p_{a_i})$ representing the motion flow field, the corresponding camera rays  $({^1P_i}, {^2P_i})$ are computed. It is important to note that for this, the pixel pair $(^1p_{a_i},^2p_{a_i})$ is converted back from panorama image to omni-image coordinates.
	
	The relation between each 3D pair $({^1P_i}, {^2P_i})$ can be described as 
	\begin{equation}
	{{^1P_i}^T} E {^2P_i} = \mathbf{0}\;,
	\label{eq:essentialMatrix}
	\end{equation}
	where matrix $E$ is called essential matrix based on epipolar geometry~\cite{hartley2003multiple}. Finally, rotation and translation (up-to-scale) between two camera poses can be extracted from $E$.
	
	%
	
	\vspace{-0.15cm}	
	\section{Implementation}
	\label{sec: implementation}
	\vspace{-0.15cm}
	The implementation to estimate the transformation between two poses $C$ from omni images $I_o$ is described in Algorithm \ref{alg:ifmi}, where the STEWENIUS-5-Points \cite{stewenius2005minimal} method is provided by the OpenGV\cite{kneip2014opengv} library.
	
	\begin{algorithm}[htbp!]
		\small
		\caption{Extended iFMI-based rotation estimation}
		\label{alg:ifmi}
		\begin{algorithmic}[1]
			\STATE \textbf{Input:} Omni images $^1I_{o}$, $^2I_{o}$;\\Noise filter thresholds $th_{pr}$, $th_{pnr}$
			\STATE Obtain panorama images $^1I_{p}$, $^2I_{p}$ of size $W \times H$\\ by cartesian-to-polar transformation 
			\STATE Extract sub-image set $^1\mathbb{A}$,  $^2\mathbb{A}$ from $^1I_{p}$ and $^2I_{p}$
			\FORALL{sub-images $^1a_i \in {^1\mathbb{A}}$, $^2a_i \in {^2\mathbb{A}}$} 
			\STATE {Compute relative motion \\ $m_i$ = iFMI($^1a_i$, $^2a_i$, $th_{pr}$, $th_{pnr}$)$= [s\;\theta\; t_x\; t_y]^\top$}
			\STATE Select pixel $^1p_{a_i} = (c_x + \delta, c_y + \delta)$, where $\delta > 0$
			\STATE Find motion pixel pair $F_i=(^1p_{a_i},^2p_{a_i})$ (Eq.~\eqref{eq:motionFlow})
			\STATE Convert $F_i$ to omni-image coordinates \\ \emph{polar-to-Cartesian}($F_i$)
			\STATE Find camera ray pair $(^1P_i, ^2P_i)$ = $\pi^{-1}(F_i)$ 
			(Eq.~\eqref{eq:cameramodel1})
			\STATE Add $(^1P_i, ^2P_i)$ to correspondences set $\mathbb{S}$
			\ENDFOR
			\STATE Transformation ${^2_1T}$ = STEWENIUS-5-Points$(\mathbb{S})$
			\STATE \textbf{Output:} ${^2_1T}$
		\end{algorithmic}
	\end{algorithm}
	
	
	\vspace{-0.15cm}	
	\section{Experiments and Results}
	\label{sec: experiments}
	\vspace{-0.15cm}
	
\begin{figure}[t]
	\setlength{\belowcaptionskip}{-10pt}

		\centering
		\subcaptionbox{Capture Device\label{fig:demo robot}}{
			\includegraphics[height=0.215\columnwidth]{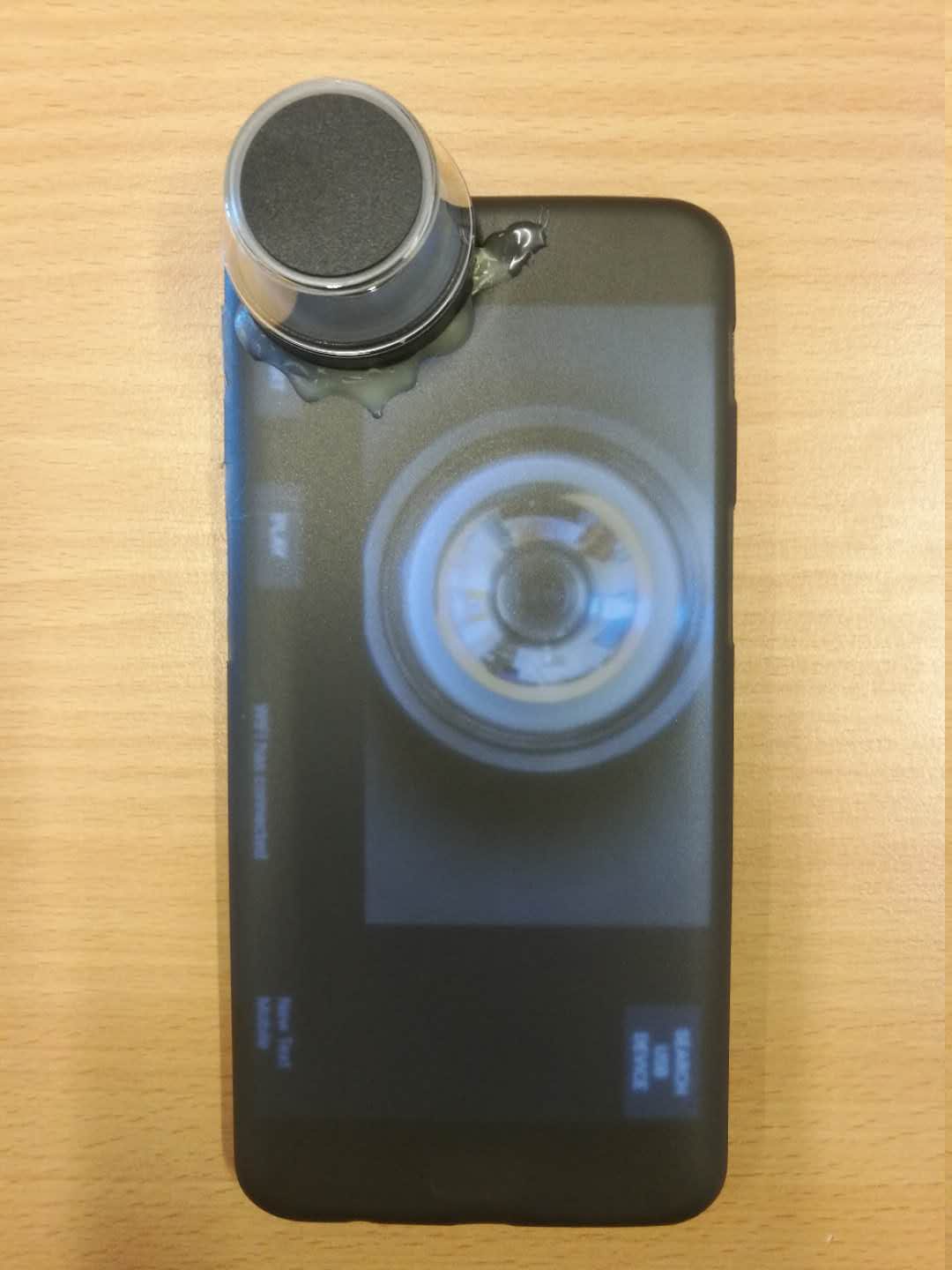} 
		} 
		\subcaptionbox{Office \\ (indoor)\label{fig:office_dataset}}{
			\includegraphics[width=0.22\linewidth]{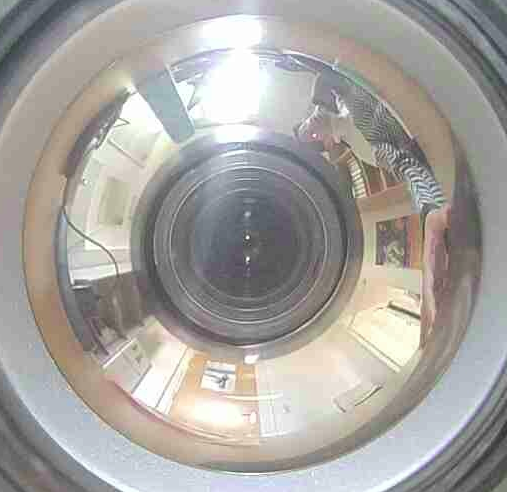}
		}
		\subcaptionbox{Lawn \\ (Outdoor)\label{fig:lawn_dataset}}{
			\includegraphics[width=0.22\linewidth]{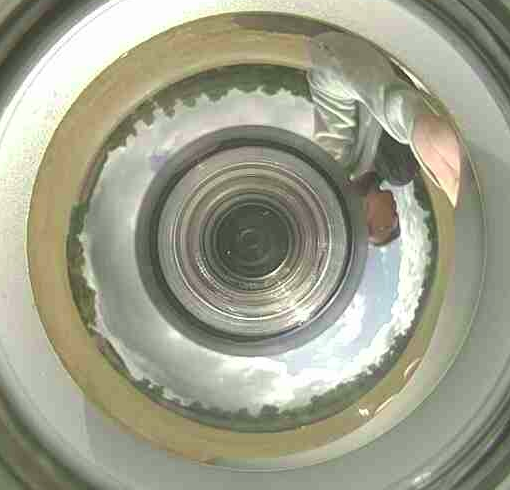}
		}
		\subcaptionbox{MPI-omni \\ ~\cite{Schoenbein2014IROS}\label{fig:libomni_dataset}}{
			\includegraphics[width=0.21\linewidth]{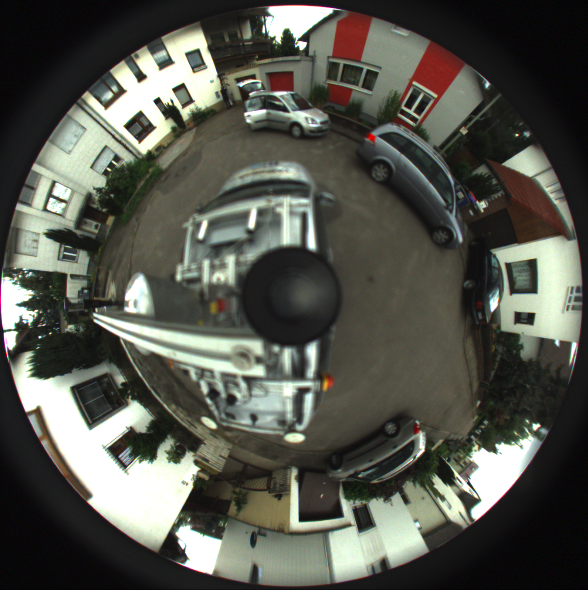}
		}
		\caption{Capture device and samples from used datasets.}
		\label{fig:dataset_samples}
	\end{figure}

	Experiments to evaluate the accuracy of the mentioned approaches in three datasets are presented here.
	The first two datasets were collected in different environments: indoor (office) and outdoor (lawn) with 500 images each.
	A smartphone equipped with an omni-lens ($\approx15$USD) was used (Fig.~\ref{fig:demo robot}) for their collection. 
	The third dataset, referred as \emph{MPI-omni} in this work, was obtained from \cite{Schoenbein2014IROS,Schoenbein2014ICRA}. 
	It contains omni-images from urban scenes collected from a driving platform; the first driving sequence of 200 frames is used to prove the universality of our approach. Ground truth for all datasets is taken from their respective Inertial Navigation Systems (INS) measurements
	and only camera orientation is compared in these experiments.
	The ranslation estimation needs further optimization techniques, because it is up-to-scale for monocular images, so this topic is out of the scope of this paper. Also, we do not compare against the original iFMI method, because it does not work with omni-images and it is not applicable when the forward-looking camera's motion has a yaw component.

	\begin{figure*}[!bht]
		\centering
		\begin{subfigure}[b]{1.0\textwidth}
			\centering
			\includegraphics[width=0.3\linewidth]{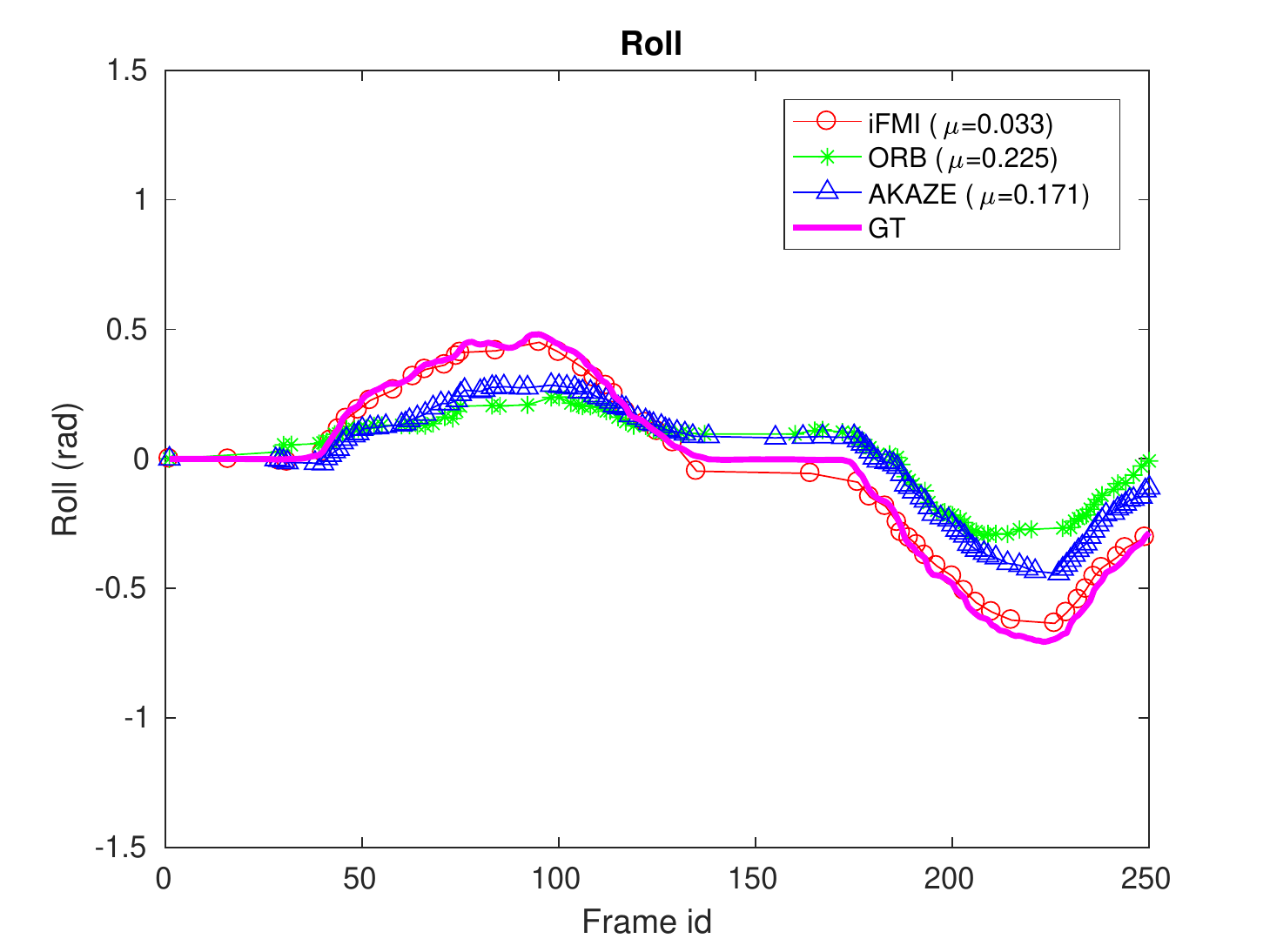}
			\includegraphics[width=0.3\linewidth]{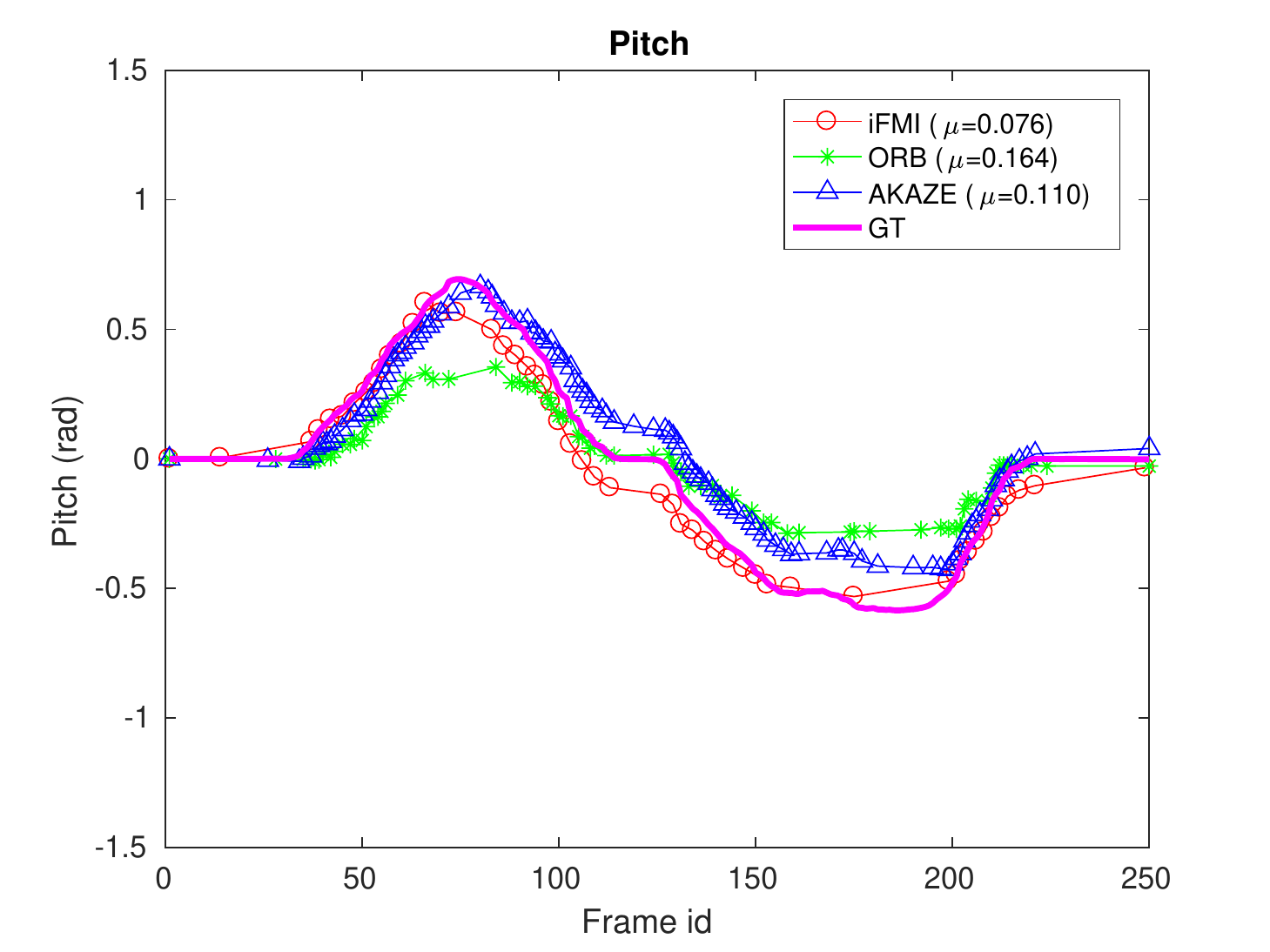}
			\includegraphics[width=0.3\linewidth]{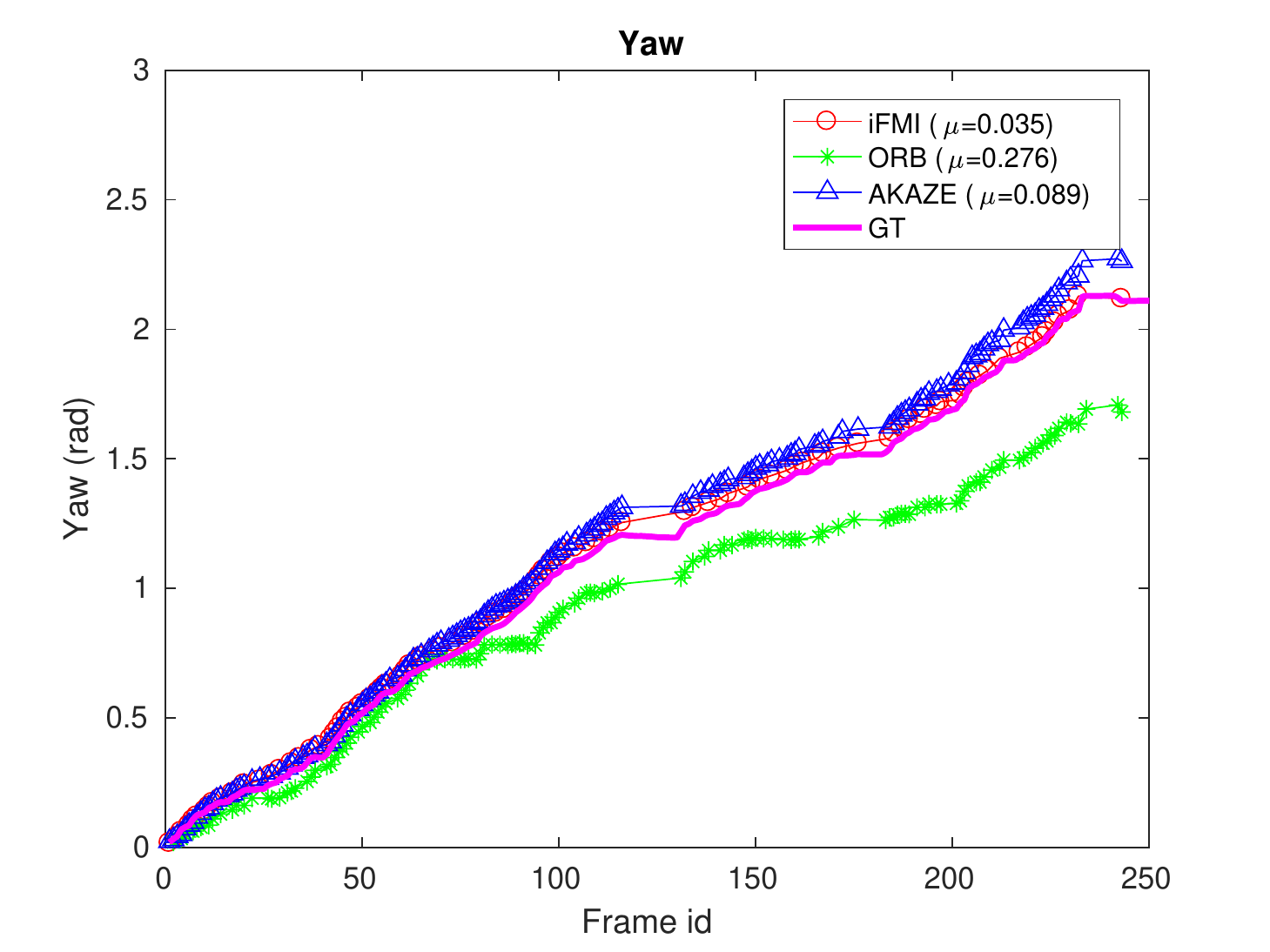}
			\caption{Office dataset}
			\label{fig:office_rot}
		\end{subfigure}
		\begin{subfigure}[b]{1.0\textwidth}
			\centering
			\includegraphics[width=0.3\linewidth]{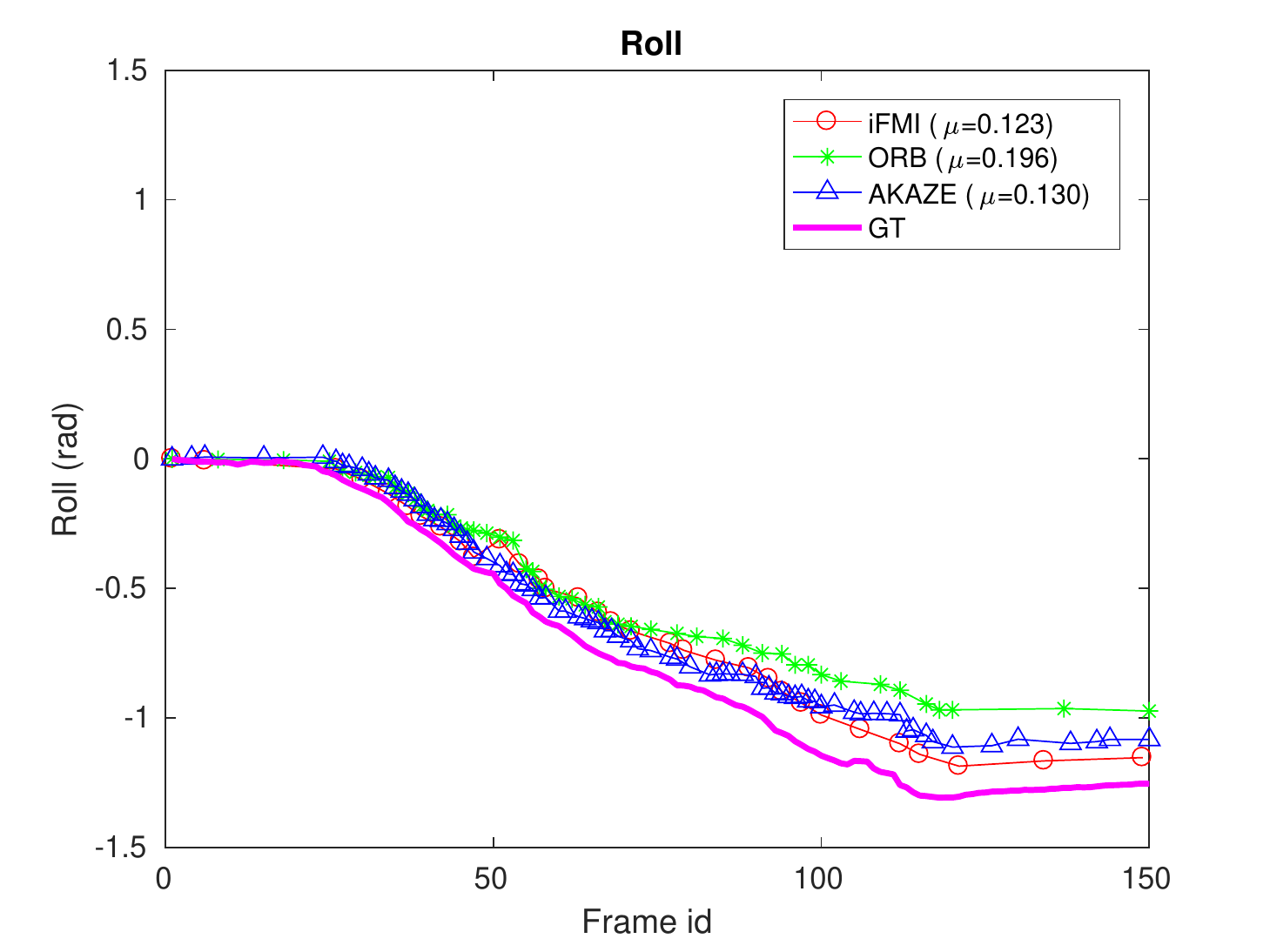}
			\includegraphics[width=0.3\linewidth]{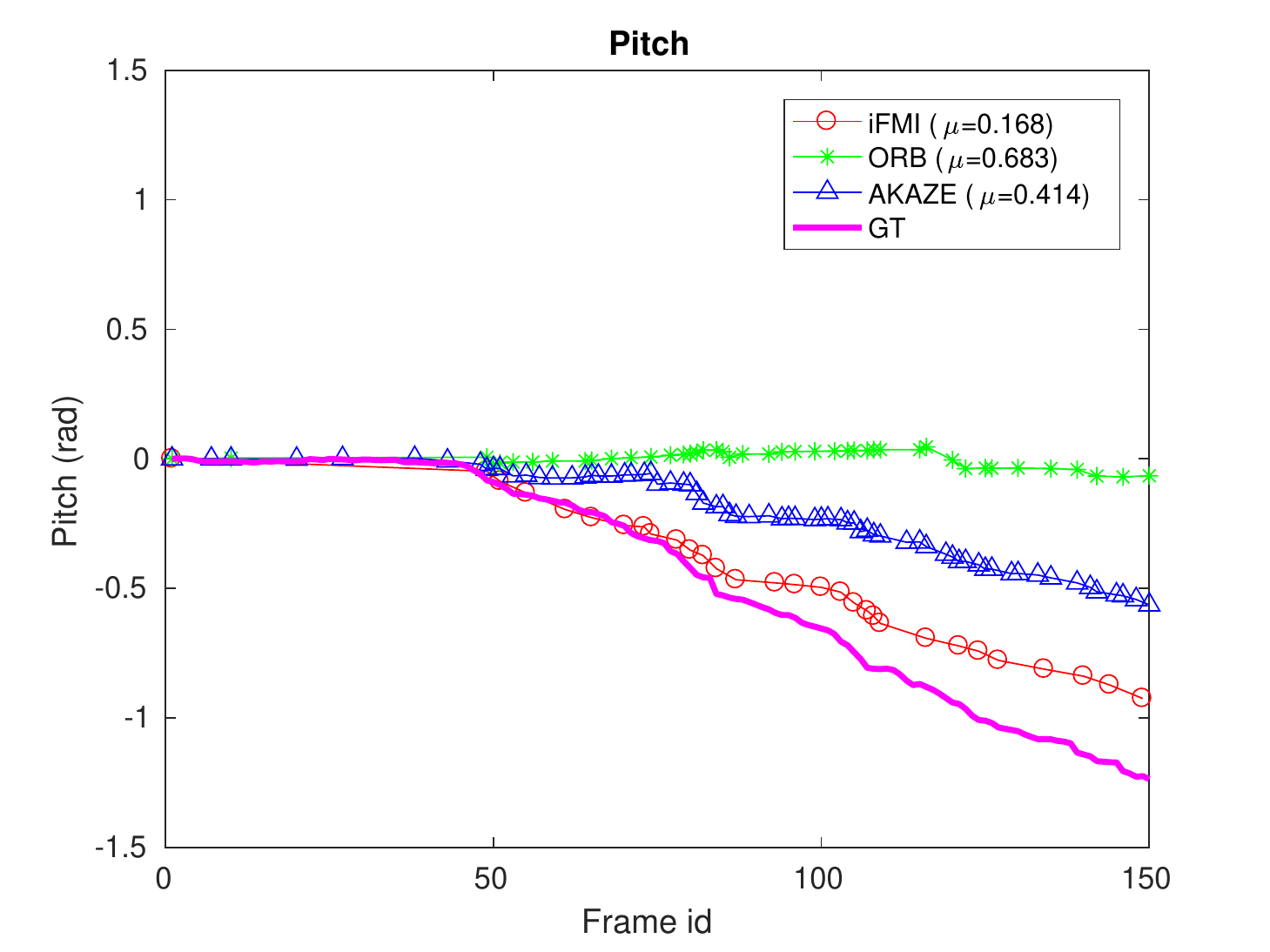}
			\includegraphics[width=0.3\linewidth]{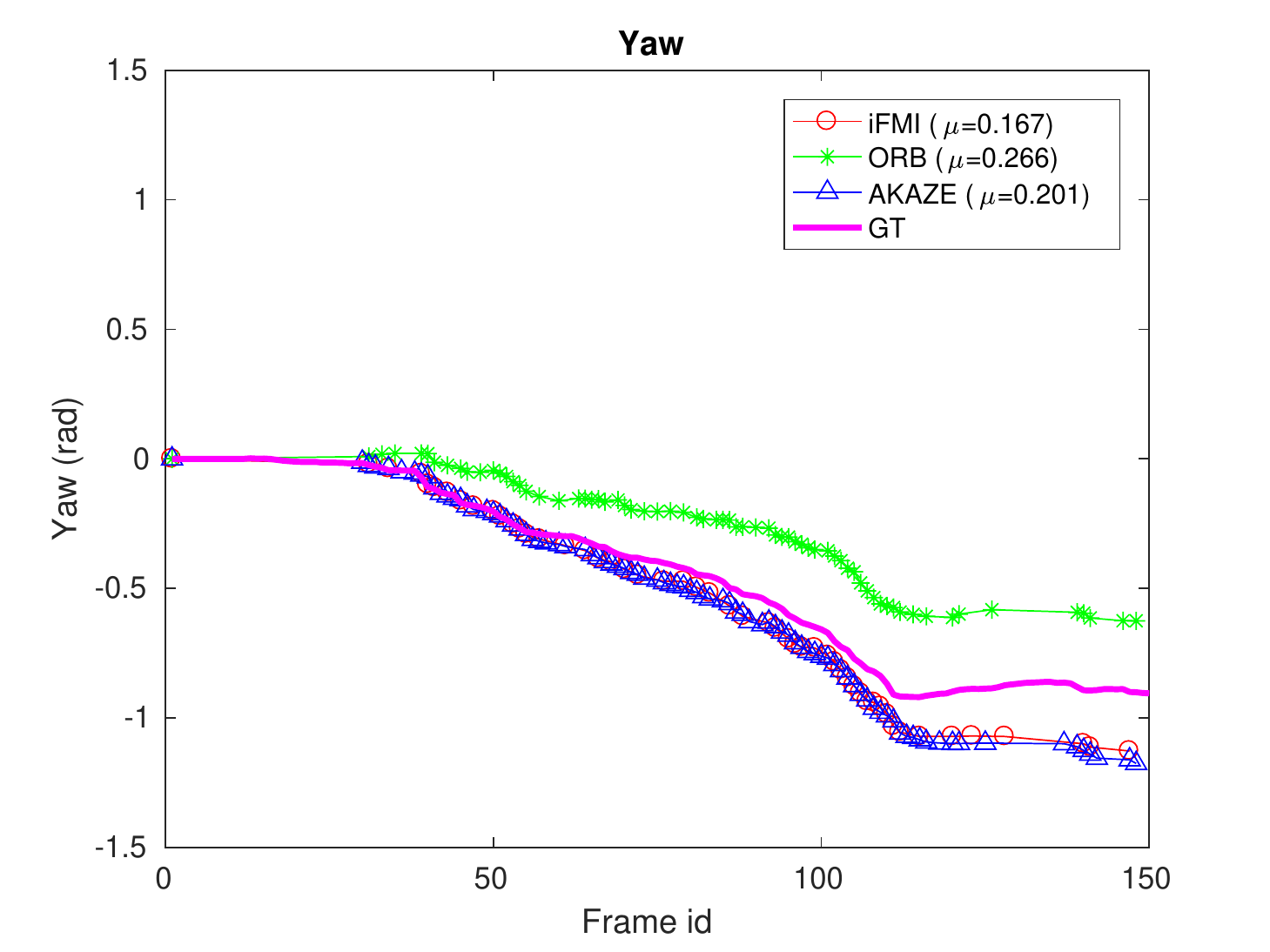}
			\caption{Lawn dataset}
			\label{fig:grass_rot}
		\end{subfigure}
		\begin{subfigure}[b]{1.0\textwidth}
			\centering
			\includegraphics[width=0.3\linewidth]{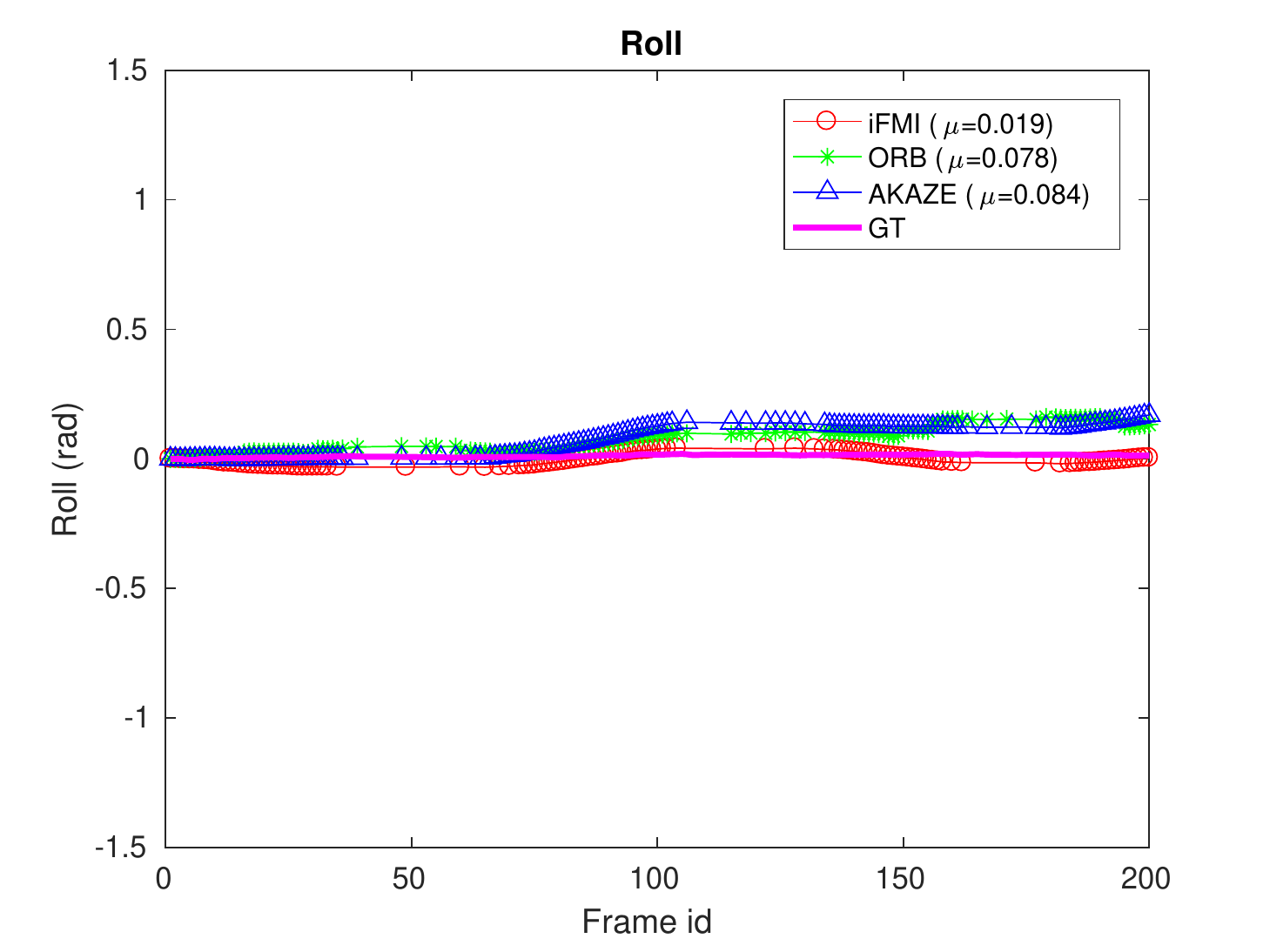}
			\includegraphics[width=0.3\linewidth]{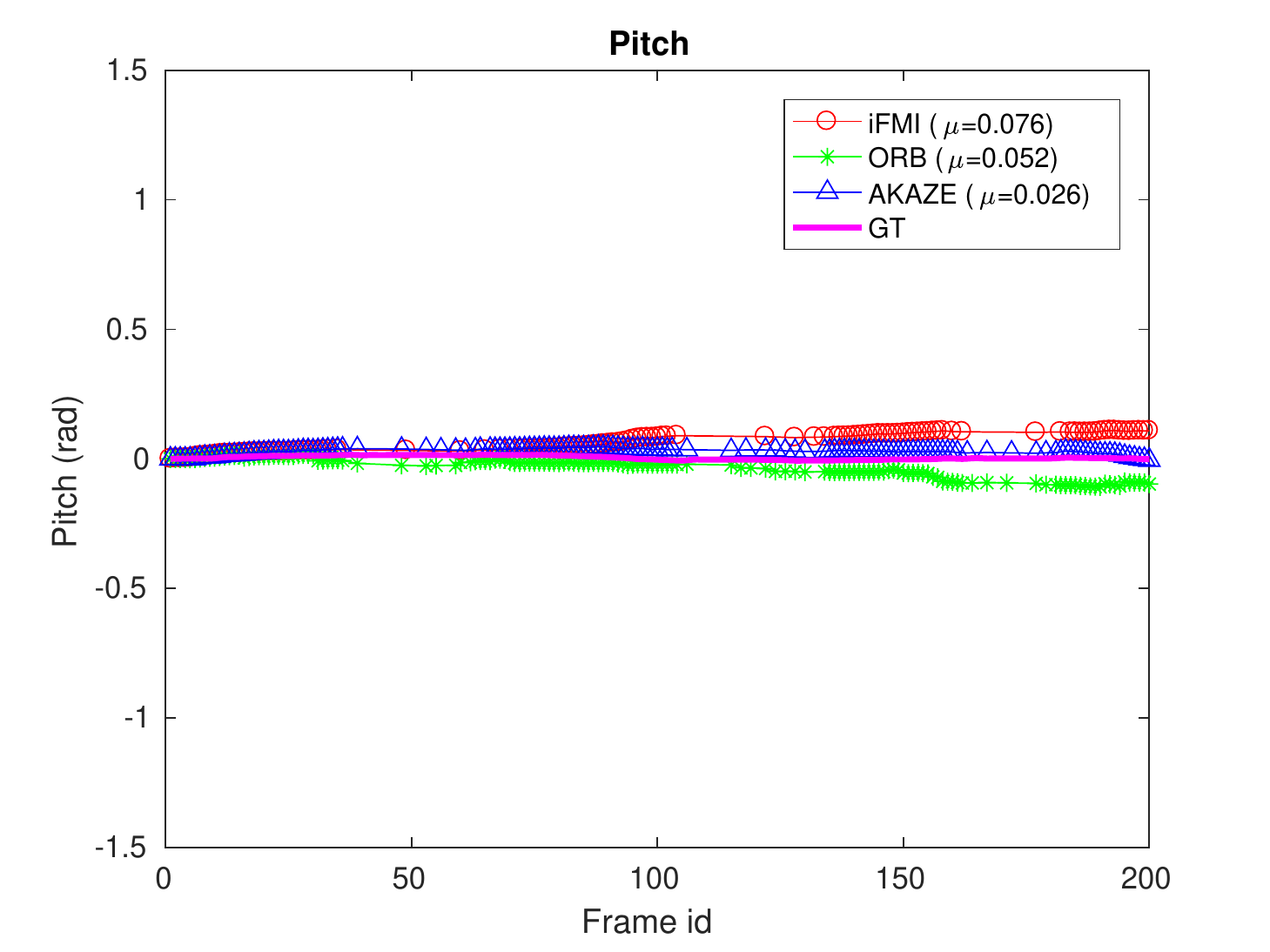}
			\includegraphics[width=0.3\linewidth]{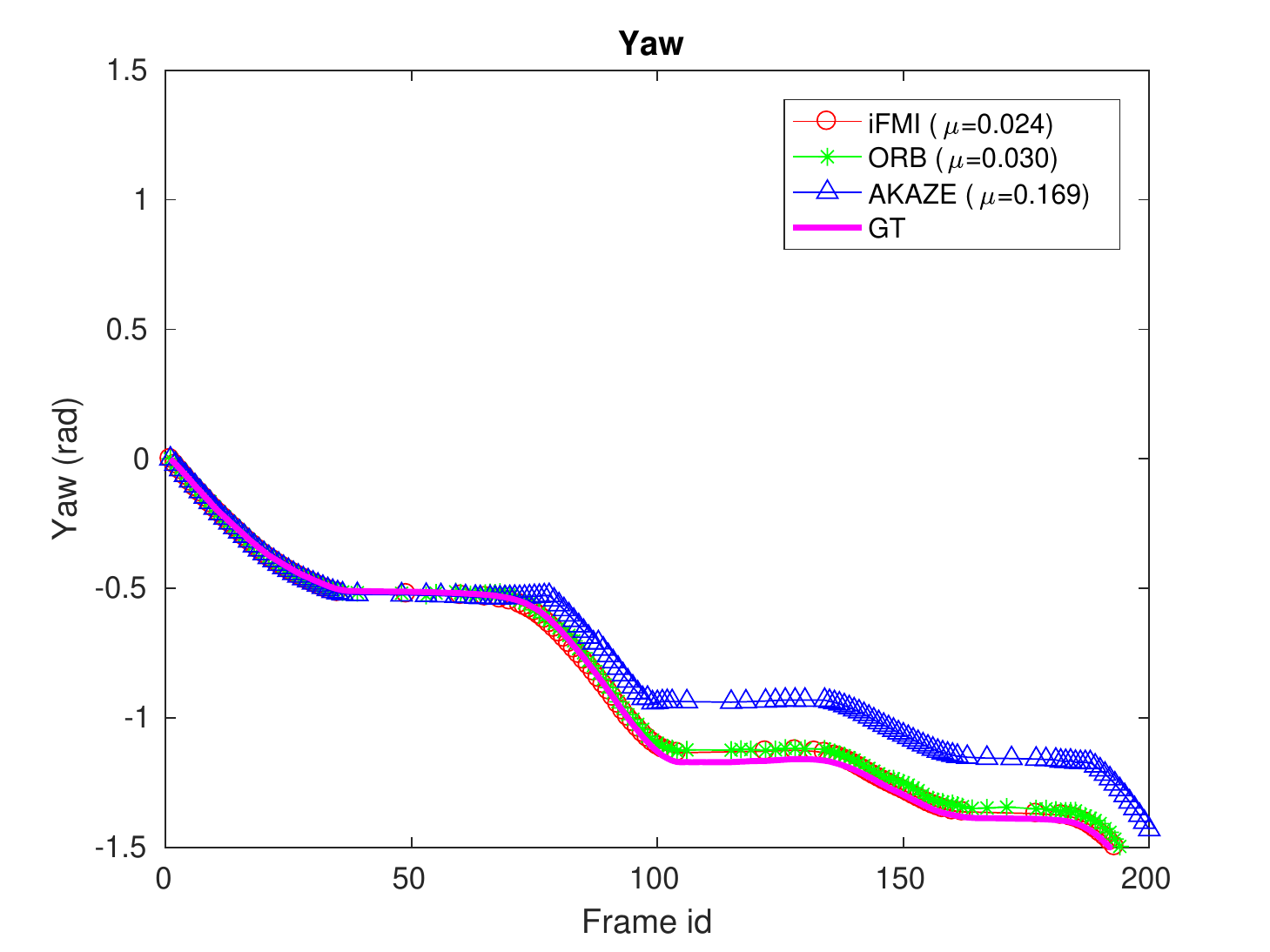}
			\caption{\emph{MPI-omni} dataset}
			\label{fig:cvlibs_rot}
		\end{subfigure}
		\caption{Orientation estimation on different datasets and absolute trajectory root mean square error $\mu$.}
		\label{fig:accuracy}
	\end{figure*}
	

	Fig. \ref{fig:accuracy} shows the orientation estimation performance for the different datasets. 
	The reported error is the absolute trajectory root mean square error, i.e., RMSE of euclidean differences. 
	In the office scenario, the results from the extended-iFMI method are closest to the ground truth, followed by AKAZE. 
	Commonly, this environment presents different objects with distinct features e.g., books, chairs, furniture, etc., that help feature registration algorithms to recover from errors between frames. 
	However, for scenarios like the lawn dataset, the images textures generate many ambiguous features that cannot be correctly matched.
	Fig. \ref{fig:grass_rot} shows that the ORB based method is unable to track pitch motion and constantly underestimates yaw. 
	Extended-iFMI still offers the best results, especially for pitch rotations.
	Finally, results on the \emph{MPI-omni} dataset in Fig. \ref{fig:cvlibs_rot} show that our approach is accurate and robust in highly dynamic environments, having the least error as well.
	
	
	\begin{table}[!b]
		\small
		\centering
		\captionsetup{justification=centering}
		\begin{tabularx}{\linewidth}{@{}szXXX@{}}
			\toprule
			~ & ~ & iFMI & ORB & AKAZE \\
			\midrule
			roll & $\epsilon[rad]$ & \textbf{0.058} $\pm$ 0.056 & 0.166 $\pm$ 0.078 & 0.128 $\pm$ 0.044\\
			pitch & $\epsilon[rad]$ & \textbf{0.107} $\pm$ 0.053 & 0.300 $\pm$ 0.337 & 0.183 $\pm$ 0.204\\
			yaw & $\epsilon[rad]$ & \textbf{0.075} $\pm$  0.080 & 0.191 $\pm$ 0.139 & 0.153 $\pm$ 0.058\\
			\midrule
			$\mu(\epsilon)$ & $[rad]$ & \textbf{0.080} $\pm$ 0.025 & 0.219 $\pm$ 0.071 & 0.155 $\pm$ 0.028\\
			\midrule
			$ \mu (t)$ & $[s]$ & 0.12 & \textbf{0.11} & 0.75\\
			\bottomrule
		\end{tabularx}
		\caption{Average orientation root mean square error (RMSE) and computation time per frame from all datasets samples.}
		\label{tab:aver_accuracy}
	\end{table}
	
	Table \ref{tab:aver_accuracy} shows the average computation time $\mu(t)$ and the average RMSE $\mu(\epsilon)$ from all three datasets measurements. ORB and extended-iFMI methods are about seven times faster than AKAZE.
	The extended-iFMI outperforms feature-based methods in roll, pitch and yaw, showing a robust performance in all types of environments.
	Its RMSE error is approximately three times smaller than that of ORB and two times smaller than that of AKAZE. 
	This is important because, it eliminates the necessity of selecting or designing a specific feature detector for certain scenarios or image distortions. 
	
	\vspace{-0.3cm}
	\section{Conclusions}
	\label{sec: conclusion}
	\vspace{-0.15cm}
	In this paper, we extend the iFMI method to calculate 3D motion between omni-images.
	Our approach is compared against two feature-based methods, ORB and AKAZE, for robustness, accuracy and run-time. 
	The experiments show that our method has the highest accuracy overall and that it is more robust against different types of environments.
	For these reasons, we believe in extending this type of spectral based approach for full 6DOF pose recovery in mobile robots.
	It is a step forward for the practical use of catadioptric omni-cameras, where accuracy and runtime are crucial.

	\bibliographystyle{IEEEbib}
	\bibliography{references}

\end{document}